\documentclass[final,3p,11pt]{elsarticle}
\usepackage{amsmath,amssymb,booktabs,array,graphicx}
\usepackage{xr-hyper}
\usepackage[hidelinks]{hyperref}
\biboptions{numbers,sort&compress}
\journal{Solar Energy}
\makeatletter\let\ps@pprintTitle\ps@plain\makeatother
\date{27 November 2026}
\newcommand{\clip}{\operatorname{clip}}
\usepackage{chngcntr}

\begin{document}
\begin{frontmatter}
\title{A Retrospective Study of Multiscale Power Residuals and Clear-sky-index Persistence for Four-hour Photovoltaic Forecasting}
\author{%
\parbox{0.94\textwidth}{\centering
Sun Ze\textsuperscript{1}, Zhou Liguo\textsuperscript{1,2,*}, Xu Yuqing\textsuperscript{1},\\
Yu Lei\textsuperscript{1}, Jiang Mingming\textsuperscript{1}, Alois Knoll\textsuperscript{2}\\[6pt]
{\small\textsuperscript{1}School of Computer Science and Technology, Huaibei Normal University, China}\\[2pt]
{\small\textsuperscript{2}Chair of Robotics, Artificial Intelligence and Real-time Systems,\\
Technical University of Munich, Germany}\\[3pt]
{\small\textsuperscript{*}Corresponding author. E-mail: zhouliguo@chnu.edu.cn}%
}}
\begin{abstract}
An hourly irradiance forecast may miss a power drop recorded by a photovoltaic meter. A four-hour forecast must then decide how much of that departure to carry into each future quarter. Confidence-weighted innovation flux transport (CIFT) measures the signed departure from a shared weather-conditioned power response at past times. It corrects the target-time response with four residual summaries and mixes this estimate with the latest daylight power-to-envelope ratio projected onto the target's solar envelope. We evaluated 16 quarter-hour leads at eight Australian plants and ten Belgian provincial aggregates in three retrospective periods. In eight-epoch matched refits, four residual scales reduced daylight capacity-normalized mean absolute error by 0.21--0.31 percentage points relative to one scale. Adding the solar-index candidate to the tested residual path reduced error by a further 0.45--0.58 points. Unrestricted forward coefficients improved on exponential attenuation in period II. The three-seed CIFT forecast reduced absolute error by 33.6--35.5\% relative to smart persistence. Pooling all 16 leads, three-seed CIFT reduced nMAE relative to LightGBM by 0.207, 0.006 and 0.025 capacity percentage points in periods I--III, and the paired interval separates in period I. The structural refits support carrying recent power information forward, with the benefit of a particular decay law varying across periods.
\end{abstract}
\end{frontmatter}

\section{Introduction}
A fall in PV output at 10:00 can improve the 10:15 forecast even if the hourly weather forecast missed the cloud. Whether the same drop should affect the 14:00 forecast depends on how the sun, local cloud cover and plant operation evolve. The meter records the combined electrical effect of cloud cover and plant operation. Coarse weather forecasts supply a future trajectory but can miss the timing of the local event.

Two familiar corrections extrapolate different quantities. A power residual carries a discrepancy in units of capacity. Clear-sky-index persistence carries a fraction of attainable power and lets the diurnal power reference change between observation and target. These corrections can agree near noon and diverge sharply on a rising or falling solar profile. Confidence-weighted innovation flux transport (CIFT) constructs both from information available under a specified issue-time convention. In the model name, confidence denotes a learned weight on measurement history, and flux denotes a bounded exchange of predicted power between neighbouring quarters. The exchange preserves the four-hour energy of that branch. The larger empirical gains came from representing recent departures and projecting the solar index.

\subsection{Relationship to existing forecasts}\label{sec:r3related}
PV forecasting methods differ in how they combine irradiance forecasts, geometry and electrical measurements \cite{anton}. Physical models such as PVWatts translate irradiance into electrical output using array and operating assumptions \cite{pvwatts}. A PV clear-sky index and statistical blends of persistence with climatology already account for recurring solar variation \cite{eng,yang}. Physics-informed irradiance persistence and learning models also address changing atmospheric attenuation \cite{part1,liuphysics}. For electrical power, the reference must accommodate plant operating range as well as solar geometry. Bacher et al.\ examine how the value of recent power and forecast weather changes with lead, and Bellinguer et al.\ integrate generation and weather measured at different time resolutions \cite{bacher,bellinguer}. Satellite cloud-motion forecasts can follow a moving cloud field between irradiance updates \cite{perez}. We consider a plant operator with power telemetry and archived gridded weather.

The closest structural comparison is Ou Yang et al.'s four-hour PV framework, which uses a physical NWP power baseline and an attention-based sequence model to predict its residual \cite{ouyang}. In CIFT the same learned solar-conditioned response supplies both the historical reference and the future baseline. The difference between measured and reference power at each valid historical quarter defines a signed, observable input. Each target quarter can combine several histories of that input with the latest power-to-reference ratio and a weather estimate when telemetry is sparse. This explicit choice of what to carry forward distinguishes the architecture. Residual correction, exponential weighting and solar-index persistence each have precedents. Because their implementation differs in inputs and training, we use this framework as a structural reference rather than a numerical benchmark on these archives.

General sequence models can use future irradiance when given target covariates \cite{vaswani,itrans,timesnet,dlinear,tsmixer}. They need not express an observed power departure separately from their hidden states. Multi-site selection and pooling address a different source of skill \cite{chen,solnet,global}; recent bias correction can also be applied to combinations of complete forecasts \cite{related}. Zhang et al.\ compared deep PV models with XGBoost and classical gradient boosting driven by target-time weather \cite{crossunetPV}. That static regression does not explicitly extrapolate a measured power departure. Here the three tree baselines also receive lagged power. We include these and neural comparators, and isolate individual choices on a common backbone. The study makes three contributions.
\begin{enumerate}
\item We define the power innovation against the same weather-conditioned response that supplies the target forecast. This gives the correction a measured sign and a scale in units of plant capacity.
\item We test two distinct extrapolations of recent power in matched refits: carrying a signed discrepancy forward across four time scales and projecting a dimensionless power ratio onto the changing solar envelope. Relaxing the forward coefficients tests the exponential assumption separately.
\item We quantify where these corrections help in a four-hour forecast by lead and by reporting scale, using the same scored targets for smart persistence, an irradiance-to-power chain, three boosted-tree methods and public neural models.
\end{enumerate}

\section{Model}
\subsection{Forecast origin and solar power reference}
\begin{figure}[tbp]
\centering
\makebox[\linewidth][c]{\includegraphics[width=1.08\linewidth]{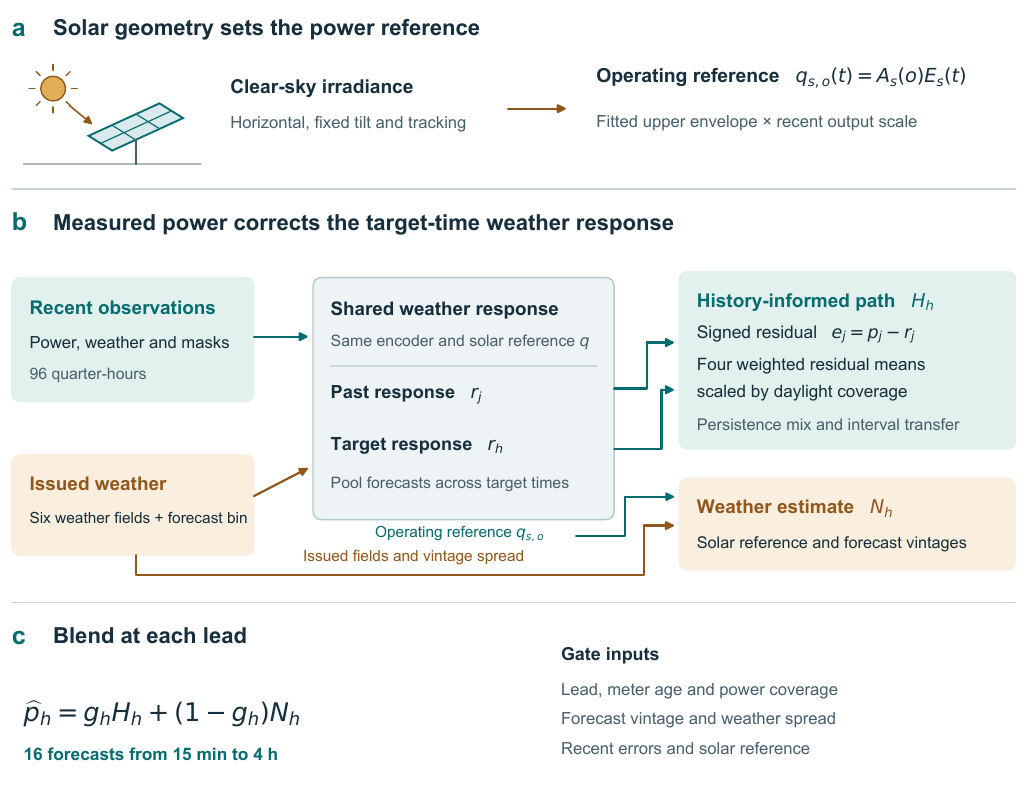}}
\caption{Information available to CIFT at issue time. (a) A site envelope $E_s$ and a recent operating scale $A_s$ define the solar reference. (b) The historical response defines the power residual. The target response supplies the base to which the weighted signed residuals are added. The weather path also receives $q$, which depends on the preceding 72 hours of measurements. (c) A lead-specific gate produces the 16 forecasts.}
\label{fig:architecture}
\end{figure}

At origin $o$, the model predicts mean power as a fraction of series capacity for sixteen intervals of length $\Delta=0.25$~h. Interval $h$ ends at $o+h\Delta$. Solar geometry is evaluated at its midpoint, where apparent solar elevation above $0^\circ$ defines the daylight indicator $d_h$. Inputs include 96 quarters of generation and weather with validity masks, future geometry and archived weather eligible at $o$. A 72-hour power history supplies the operating scale. Figure~\ref{fig:architecture} shows where measurements enter the forecast.

For series $s$ at its representative coordinates and target time $t$, let $b_s(t)$ contain three clear-sky irradiance proxies for horizontal, equator-facing fixed-tilt (tilt equal to absolute latitude), and north--south single-axis tracking surfaces limited to $60^\circ$ without backtracking. The proxies use solar position and a Kasten--Young air-mass estimate and a Meinel-type irradiance approximation \cite{kasten,meinel}. Each represents a possible diurnal shape. A site-specific upper power envelope is
\begin{equation}
 E_s(t)=\min\{c_s,\,w_s^\top b_s(t)\},\quad w_s\ge0,\quad 0<c_s<1.3 .
 \label{eq:envelope}
\end{equation}
We fit the weights and clipping level to valid training observations with solar elevation above $5^\circ$ by minimizing the pinball loss at quantile 0.97. These weights can describe a mixture of diurnal profiles even when the orientation of the actual modules is unknown.

The 0.97 quantile tracks the upper portion of training output while limiting the influence of individual peaks. The bound $c_s<1.3$ limits the fitted site envelope, while $P_{\max}=1.25$ caps the final capacity-normalized power forecast. The two bounds apply at different stages. To accommodate recent changes in attainable output, we form $A_s(o)$ from the 95th percentile of $p_s(t)/E_s(t)$ among valid intervals in the previous 72 hours with $E_s(t)\ge0.30$. The factor is clipped to $[0.02,1.20]$ and defaults to 1 when fewer than eight observations qualify. A simulated telemetry gap removes the same observations from this estimate as from the power input. The resulting reference $q_{s,o}(t)=A_s(o)E_s(t)$ can follow a temporary reduction in production. A prolonged cloudy period can also depress the ratio, so $A_s$ reflects recent operating level rather than a direct availability measure.

The 0.95 quantile estimates recent attainable output. The 0.30 cutoff excludes small denominators near the horizon. These thresholds were held fixed across series to keep the comparison uniform.

\subsection{A shared response before and after issue time}
The weather encoder first pools the 96 historical and 16 future weather vectors available at issue time. Each target's solar geometry queries all 112 positions. Each field pools only over positions where it is present and receives a bounded solar-conditioned adjustment. A propagation layer then maps the resulting sixteen vectors across target midpoints. Its weights depend on the distance between the two midpoints and on a learned decay kernel. All sixteen target-weather vectors are assigned the issue time $o$ in this experiment. Their source age in the propagation kernel is therefore zero. The age--lead interaction is consequently inactive in this implementation. Supplementary Section~\ref{Sup-sec:jgd-equations} specifies the kernel, the attention and the distinct archive-bin input used elsewhere.

The shared encoder maps geometry, weather, field masks, an age input and an eight-dimensional series embedding to 64 features. Historical intervals use the weather aligned to each interval. At future intervals the encoder receives pooled weather. Its state receives one projection of the propagated weather and another of sixteen residual statistics and four local shortwave and cloud statistics. Let $z_t$ be the resulting state for the corresponding interval. The same linear readout $\ell_\theta$ gives
\begin{equation}
 r_t = d_t P_{\max}\,
 \sigma\!\left(\ell_\theta(z_t)+
   \operatorname{logit}\{\clip(q_{s,o}(t),0.002,0.98)\}\right),
 \quad P_{\max}=1.25 .
 \label{eq:response}
\end{equation}
The stored geometry channel is $q_{s,o}(t)/0.8$. The readout multiplies this channel by 0.8 before clipping. In daylight, setting the learned logit to zero gives 1.25 times the clipped reference. For example, $q=0.4$ gives $r=0.5$, and all $q\ge0.98$ give $r=1.225$. The fitted model uses the envelope as a log-odds offset with this scale. The past readout defines the innovations below. The future readout already depends on their summary before the explicit additive correction.

\subsection{Multiscale memory of recent power residuals}
For history position $j=1,\ldots,96$ in chronological order, let $m_j$ indicate a valid daylight measurement, $e_j=p_j-r_j$ on that support and $a_j=(96-j+1/2)\Delta$ its midpoint age. Four learned time scales $\tau_k=\clip(\exp\theta_k,0.125,168)$~h give
\begin{equation}
 \mu_k=\frac{\sum_j m_j\exp(-a_j/\tau_k)e_j}
                 {\sum_j m_j\exp(-a_j/\tau_k)} .
\end{equation}
Each denominator has a floor of $10^{-6}$. The summary concatenates four weighted means, four weighted adjacent differences, four standard deviations and four coverage fractions, in that order. Coverage $c_k$ divides observed daylight kernel mass by all daylight kernel mass. Only consecutive quarters with both observations valid enter a difference. Supplementary Section~\ref{Sup-sec:jgd-equations} gives the weights and empty-history conventions.

At lead $h$, a network reads $z_h$, the summary and $h\Delta/4$ to produce four softmax weights $\pi_{hk}$ and four softplus rates $\lambda_{hk}$. At the interval end the history-adjusted estimate is
\begin{equation}
 u_h=d_h\,\clip\!\left(r_h+\sum_k\pi_{hk}c_k\mu_k
             \exp\left\{-\Delta\sum_{i=1}^h\lambda_{ik}\right\},
             0,P_{\max}\right).
 \label{eq:memory}
\end{equation}
For each scale, the exponential multiplier is non-increasing with lead. The mixing weights can increase with lead. The future response also reads the memory summary, and one persistence candidate uses the latest discrepancy directly. No monotonicity follows for the mixed correction or the final forecast. In the selected configuration the historical readout is detached when it supplies $r_j$ for the innovation. Forward predictions are unchanged by this operation for a given set of weights. Training can still update the shared response through its future readout. We evaluate this optimization choice separately from pooling across sites.

\subsection{Persistence, redistribution and the weather path}\label{sec:mix}
Let $j^\star$ index the last valid daylight observation, and write $q_h$ for the solar reference at the target quarter. The history path mixes $u_h$ with two daylight-masked candidates clipped to $[0,P_{\max}]$. The first extrapolates $q_h$ using $p_{j^\star}/\max(q_{j^\star},0.05)$ clipped to $[0,1.5]$. The second uses $u_h+e_{j^\star}$. The last innovation therefore enters the weighted history and this candidate a second time. The softmax reads $z_h$, $h\Delta/4$ and the age of the last valid daylight quarter measured from its start. With no such quarter it assigns all weight to $u_h$.

Let $v_h$ be the resulting mixture and $b_h=\min(v_h,P_{\max}-v_h)$. Between adjacent quarters the signed transfer is $F_h=0.45\min(b_h,b_{h+1})\tanh(\eta_h)$, with $F_0=F_{16}=0$. The path returns $H_h=v_h+F_{h-1}-F_h$. The differences telescope, so $\Delta\sum_h H_h=\Delta\sum_h v_h$ for the history path alone. Together the two edges at an interval can use at most $0.90b_h$, preserving the power bounds and zero night output.

In parallel, the weather path uses the unpooled six-field forecast, its archive-bin index and summaries over eligible vintages. With stored solar channel $C_h=q_h/0.8$ and weather features $x_h$ (\ref{Sup-sec:jgd-equations}), it gives $N_h=d_h\,\clip(1.5C_h\,\sigma(f_N(x_h)),0,1.2)$. Power from 24--72 hours before issue can still affect $C_h$ when the latest day is masked. The final forecast is
\begin{equation}
 \hat p_h=g_h H_h+(1-g_h)N_h,\qquad 0<g_h<1 .
\end{equation}
Here confidence denotes the weight given to a correction as a function of input coverage and forecast conditions, not a calibrated probability. The sigmoid gate reads lead, telemetry age and coverage, forecast-bin index, shortwave and cloud disagreement across vintages, recent forecast discrepancies and $C_h$. Its lead input runs from 0 to $15/16$. The gate is fitted from power errors. For these public series each branch is already daylight-masked and bounded by $P_{\max}$. The inference wrapper clips the mixture and reapplies daylight, which leaves finite in-range predictions unchanged.

\section{Data and evaluation}
\subsection{Generation, weather and issue times}
Lettered tables, figures and sections are in the separate Supplementary material.
We use eight Australian utility-scale generation series (AEMO) and ten Belgian provincial aggregates (Elia). Each Australian target is a mean of three five-minute readings, normalized by registered capacity and screened for dispatch limitation. Belgian targets are Elia's measured and upscaled provincial estimates, expressed as capacity factors and derived from Elia's ODS032 and ODS209 archives \cite{aemo,elia,elia209}. For Belgian weather we used one approximate geographic centroid per province, so a province shares a single weather trajectory across its installed capacity. The Supplementary material identifies all series (Table~\ref{Sup-tab:ids}) and checks the October transition between Belgian archives (Section~\ref{Sup-sec:audit}).

For period III we counted the AEMO dispatch flags among daylight origin--lead pairs before quality filtering. The flagged shares were 4.85\% at Beryl, 44.57\% at Emerald, 0.50\% at Glenrowan, 30.81\% at Haughton, 8.81\% at Moura, 0.33\% at New England, 18.04\% at Sunraysia and 1.22\% at Western Downs. Multiple forecasts can cover the same delivery quarter. At Emerald the screen retained 3,313 of 5,977 pairs. These percentages describe evaluation exposure to dispatch limits under the issue schedule.

Weather inputs come from the Open-Meteo archive of ECMWF IFS forecasts on a $0.25^\circ$ grid \cite{openmeteo}. All four quarters of a weather hour share its mean shortwave flux. The primary fields are reconstructed from fixed-lead archive bins under an assumed eight-hour publication delay. Meter histories use the quarter ending at issue time under ordinary availability. The dispatch screen also relies on archived records. During period III, Belgian nMAE for CIFT was 3.658\% before the archive transition and 2.697\% afterwards. The two sources cover different dates, so this difference reflects weather variation together with any change of measurement. A live deployment would align these records on contemporaneous weather, meter and dispatch receipts.

\subsection{Training and temporal evaluation}\label{sec:r3split}
We fit on hourly origins from 10 October 2025 to 31 May 2026 and choose configurations using 1--14 June. The original neural training pool contains 64,559 qualifying origins. All three tree baselines use a four-hour boundary purge, retaining 64,499. The original validation pool contains 4,385 origins. Table~\ref{tab:windows} lists the later periods at the same 18 series. Evaluation origins are hourly from 05:00 to 17:00 local standard time (UTC+10 in Australia, UTC+1 in Belgium). All three periods had been inspected in earlier development. This replay measures temporal transfer at known sites. At early and late issue hours the scored leads change with sunrise, sunset and latitude. A Belgian province and an Australian plant each contribute one series to the mean.
\begin{table}[tbp]
\centering
\caption{Retrospective evaluation periods in 2026. Counts refer to all 18 series. Targets count valid daylight origin--lead pairs after masking.}
\label{tab:windows}
\begin{tabular}{@{}llrrr@{}}
\toprule
Period & Local origin dates & Days & Origins & Targets\\
\midrule
I   & 15 July--30 August       & 47 & 10,998 & 157,622\\
II  & 1 September--18 October  & 48 & 11,232 & 150,228\\
III & 19 October--19 November  & 32 &  7,488 &  90,135\\
\bottomrule
\end{tabular}
\end{table}

\subsection{An irradiance-to-power baseline}
We converted the same issued hourly global horizontal irradiance and air temperature supplied to CIFT into capacity-normalized AC power. At each target-quarter midpoint we calculated solar position, used the Erbs decomposition \cite{erbsR7} to obtain direct and diffuse components and transposed them to a fixed equator-facing plane tilted at the absolute latitude. We used an isotropic sky and albedo 0.2. The PVWatts DC and inverter equations were evaluated with pvlib 0.13.1 \cite{pvwatts,pvlibR7}. For temperature we used $T_{\rm cell}=T_{\rm air}+25G_{\rm POA}/800$, with irradiance in W\,m$^{-2}$ and temperature in $^\circ$C. The DC power coefficient was $-0.004$ per $^\circ$C. We assigned one unit of DC nameplate and one unit of nominal AC output per unit of registered plant or monitored provincial capacity, used a nominal inverter efficiency of 0.96 and applied no additional loss factor. Output is zero below the horizon. The coefficients were fixed before scoring this baseline. The conversion fixes one representative plane and one nominal rating per series, in place of measured array orientations, DC oversizing and provincial system mixtures. This generic conversion uses no recent power observations or fitted site coefficients.

\subsection{Training budgets and matched comparisons}
\label{sec:fair-design-r4}

CIFT and the full-budget TSMixer receive the same inputs at issue time. Both see a 96-quarter power and weather history with validity flags, 16 quarters of solar geometry and forecast weather, the availability-adjusted solar envelope, weather age, series identity and six weather fields in seven forecast-vintage bins with availability masks. TSMixer projects historical geometry, age and vintages into its 33 history channels and supplies the future covariates to its output head. XGBoost and classical gradient boosting (GBDT) \cite{xgbR5,friedmanR5} share the same engineered matrix of power lags, envelope ratios, historical weather summaries and lead-specific forecasts. For these two methods, series identity uses 18 indicator columns and missing values use a sentinel below the observed range. LightGBM uses the same underlying predictors, a categorical series identifier and native missing-value routing.

The CIFT search crossed site-specific or pooled fitting, gradient flow or detachment at the historical readout, and ordinary or outage-augmented training. Each of the eight configurations used seeds 20261016, 20261017 and 20261018. We averaged the three forecasts before scoring each configuration on ordinary June telemetry. The smallest June nMAE was 4.369\% for the pooled, detached model with ordinary training, and its outage-augmented counterpart scored 4.377\%. These two configurations were the only ones within 0.01 percentage points of the minimum, and the prespecified ordering placed the augmented fit first.

The full-budget comparison screened AdamW learning rates $10^{-3}$ and $3\times10^{-4}$ on the first seed for CIFT and for each of two TSMixer arms. June selected $10^{-3}$ for CIFT and $3\times10^{-4}$ for both TSMixer arms. All used 64,559 origins, batches of 32 (2,018 updates per epoch), a 30-epoch cosine schedule and patience of eight epochs. Training minimized absolute error over all valid labels, including night. For the augmented fits, half the training windows in expectation retained ordinary telemetry. The remainder masked the latest 4, 8, 16, 32 or 96 power quarters with equal probability and recomputed envelope availability. Rate and checkpoint selection used mean daylight MAE over ordinary and 16-quarter masked June inputs, weighting series equally. Ordinary TSMixer retained the observed telemetry masks throughout and used ordinary June MAE alone.

Each tree family used ordinary telemetry, the same 64,499 purged training origins and a separate absolute-error fit per lead. June selection used 4,385 origins and gave each series equal weight. For XGBoost and classical GBDT we fixed depth at 4, learning rate at 0.08, row sampling at 0.8 and the candidate feature fraction at 0.5 per split. Our earlier search ended at 512 rounds with validation error still declining. We extended each trajectory under a rule fixed before the extension. Every 16 rounds we scored the model on June. From round 640 onward, eight consecutive checks without a new minimum ended training, and the earliest minimum supplied the final prefix. The full series of checks and the retained and stopping rounds are in Supplementary Section~\ref{Sup-sec:r5trees} and the source package.

We also included an archived LightGBM comparator \cite{keLightGBM}. For each lead it searched four combinations of learning rate (0.04 or 0.08) and maximum leaf count (15 or 31), with patience of 40 rounds and a limit of 3,000. All three tree families ran on CPUs of the same host. Their selected June fits were applied directly to periods I--III. The comparisons align issue-time information and scoring. Both neural and tree entries use the same inputs at issue time; the neural entries add three-seed averaging and outage augmentation, and each tree family contributes a single fit per lead.

\subsection{Metric and uncertainty}\label{sec:comparators}
Following verification practice for deterministic solar forecasts \cite{yangverification}, we report absolute, squared and signed errors and identify the reference used for skill. For normalized target $p$ and forecast $\hat p$, let $\mathcal V_s$ contain valid daylight origin--lead pairs for series $s$, and let $S$ be the number of series ($S=18$ overall). The headline statistic in percent of capacity is
\begin{equation}
 \mathrm{nMAE}=\frac{100}{S}\sum_{s=1}^{S}
    \frac{1}{|\mathcal V_s|}\sum_{(o,h)\in\mathcal V_s}
         |\hat p_{s,o,h}-p_{s,o,h}| .
\end{equation}
Predictions are bounded to $[0,1.25]$ and scored with the same daylight and label masks. The metric gives each series equal weight. It does not give extra weight to large plants or to a region with more valid observations per series. For the principal paired intervals we resample circular blocks of three consecutive local origin dates, using 2,000 draws and the same dates across sites, models and leads. In the sensitivity analysis we also resample single local origin dates 2,000 times within each period, retaining all sites and leads on each sampled date. Percentile intervals use the paired difference in site-equal nMAE. This preserves within-day dependence, and the one- and seven-day blocks in Supplementary Table~\ref{Sup-tab:blocks} cover regimes that persist across days. The intervals are conditional on these sites, fitted models, periods and the validation selection. The current comparison aligns XGBoost, GBDT and LightGBM with frozen CIFT forecasts by series, issue time and lead, checking the target values and masks. Figure~\ref{fig:leads} also retains the full-budget ordinary-telemetry TSMixer. Supplementary Table~\ref{Sup-tab:blocks} gives the corresponding one- and seven-day results for the three tree comparisons. Each draw uses the same blocks across series and leads, truncates the last block to the original period length, and recomputes each series numerator and denominator before averaging. Intervals refer to individual contrasts, and the lead profiles report pointwise bands rather than a simultaneous all-lead test.

\section{Results}
\subsection{Accuracy by forecast lead}
For each individual CIFT fit we computed absolute errors before averaging across seeds. Each tree used one fit per lead. Intervals favoured CIFT against XGBoost and GBDT in period I, and the remaining intervals spanned zero. Table~\ref{Sup-tab:individual} gives all nine effect sizes and their three-day block intervals. These intervals describe variation over forecast dates conditional on the fits. CIFT has 35,162 trainable parameters per fit.

The three-seed CIFT forecast had the lowest all-lead nMAE in each period (Table~\ref{tab:overall}). LightGBM had the lowest nMAE of the three tree families. CIFT's advantage over LightGBM was 0.207 capacity percentage points in I (95\% paired interval 0.112--0.305). In II and III the gaps were 0.006 and 0.025 points, with intervals that spanned zero. For nRMSE, which weights large misses more heavily, CIFT had the lowest error in I and LightGBM in II and III. The all-lead nMAE contrasts for CIFT versus ordinary-telemetry TSMixer were statistically comparable in every period.

\begin{figure}[tbp]
\centering
\makebox[\linewidth][c]{\includegraphics[width=1.08\linewidth]{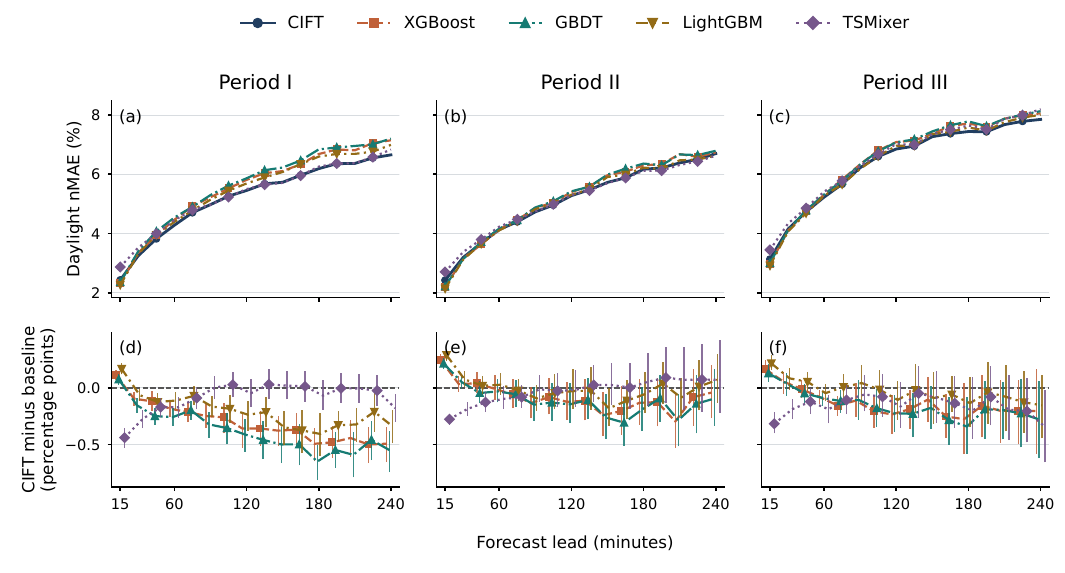}}
\caption{Forecast error by lead on the same daylight targets in 18 public series. The upper panels give capacity-normalized MAE. The lower panels show CIFT minus each comparator in percentage points, with pointwise 95\% intervals from 2,000 paired three-day block resamples. Intervals are offset horizontally within each lead for legibility. Negative values favour CIFT. TSMixer uses the full training budget and ordinary telemetry. The first forecast is for the quarter ending 15 minutes after issue.}
\label{fig:leads}
\end{figure}

The three tree families had lower error at 15 minutes in each period; by 240 minutes CIFT had lower error than all three in period I, and the period-II and period-III intervals spanned zero (Figure~\ref{fig:leads} and Supplementary Table~\ref{Sup-tab:r5contrasts}).
\begin{table}[tbp]
\centering\small\setlength{\tabcolsep}{3.5pt}
\caption{Daylight errors as percentages of capacity. Each period pools 16 leads within each series and weights all 18 series equally. All models use the same valid targets. Neural forecasts in the upper panel average three predictions; trees use one fit per lead. The lower panel averages the individual nMAEs after scoring. Bold indicates the lowest error in the upper panel within each period and metric.}
\label{tab:overall}
\begin{tabular}{@{}lrrrrrr@{}}\toprule
Forecast & \multicolumn{2}{c}{I} & \multicolumn{2}{c}{II} & \multicolumn{2}{c}{III} \\
\cmidrule(lr){2-3}\cmidrule(lr){4-5}\cmidrule(l){6-7}
 & nMAE & nRMSE & nMAE & nRMSE & nMAE & nRMSE \\
\midrule
Generic PVWatts chain & 12.420 & 16.432 & 14.220 & 18.664 & 13.323 & 18.457\\
Smart persistence & 8.066 & 12.782 & 7.622 & 12.368 & 9.591 & 14.255\\
June persistence--climatology blend & 7.862 & 11.826 & 7.862 & 11.764 & 9.377 & 13.180\\
DLinear & 6.630 & 9.796 & 6.681 & 10.111 & 7.747 & 10.974\\
TimesNet & 6.699 & 10.022 & 6.198 & 9.481 & 7.542 & 10.692\\
iTransformer & 6.132 & 9.300 & 5.989 & 9.168 & 7.258 & 10.376\\
\shortstack[l]{Full-budget TSMixer,\\ordinary telemetry} & 5.289 & 8.411 & 5.094 & 8.244 & 6.496 & 9.762\\
XGBoost & 5.504 & 8.677 & 5.140 & 8.199 & 6.467 & 9.565\\
GBDT & 5.582 & 8.757 & 5.168 & 8.237 & 6.488 & 9.588\\
LightGBM & 5.410 & 8.628 & 5.069 & \textbf{8.171} & 6.361 & \textbf{9.517}\\
CIFT & \textbf{5.203} & \textbf{8.395} & \textbf{5.062} & 8.297 & \textbf{6.336} & 9.673\\
\midrule
\multicolumn{7}{@{}l}{Mean nMAE across three individual fits}\\
CIFT & 5.324 & & 5.157 & & 6.431 & \\
TSMixer, ordinary telemetry & 5.405 & & 5.201 & & 6.594 & \\
\bottomrule\end{tabular}
\end{table}

Using a June-weighted blend of smart persistence and recent climatology as the reference, CIFT reduced all-lead nMAE by 33.8\%, 35.6\% and 32.4\% in I--III (Table~\ref{Sup-tab:r8sensitivity}). The blend reduced nRMSE relative to smart persistence in each period but increased nMAE in II, where its June squared-error weights transferred less well under absolute loss.

The generic PVWatts forecast had all-lead nMAE of 12.420, 14.220 and 13.323\%. Its largest regional signed bias was in Belgium, where it overpredicted by 8.85--9.40 points across periods. One representative plane, a nominal DC rating and a fixed conversion efficiency define this generic estimate, and part of its gap to CIFT reflects the recent power that CIFT receives. Supplementary Table~\ref{Sup-tab:physical} gives the paired comparison.

\subsection{Plant and provincial forecasts}
Table~\ref{tab:region-main} separates the eight Australian plants and ten Belgian provinces. CIFT minus LightGBM in plant-level nMAE was -0.202, -0.114, -0.038 points in I--III. The corresponding differences for the Belgian provinces were -0.211, +0.080, -0.014. The signed errors in the same table distinguish persistent over- or underprediction from absolute error. Supplementary Table~\ref{Sup-tab:r8sensitivity} shows how the comparison changes when each series is normalized by its mean daylight power.

\begin{table}[tbp]\centering\small
\caption{Daylight nMAE / signed bias in capacity percentage points. Forecast minus measured power defines the sign. Rows weight the eight plants or ten provinces equally. CIFT averages three forecasts; each tree uses one fit per lead.}\label{tab:region-main}
\begin{tabular}{@{}llccc@{}}\toprule
Archive & Forecast & I & II & III\\\midrule
AEMO & CIFT & 7.594 / +0.882 & 7.457 / +1.311 & 10.604 / +1.547\\
AEMO & XGBoost & 7.912 / +0.502 & 7.690 / +1.229 & 10.776 / +1.456\\
AEMO & GBDT & 8.046 / +0.432 & 7.737 / +1.223 & 10.826 / +1.449\\
AEMO & LightGBM & 7.796 / +0.377 & 7.571 / +1.169 & 10.642 / +1.427\\
\midrule
Elia & CIFT & 3.291 / +0.694 & 3.147 / -1.292 & 2.921 / -0.681\\
Elia & XGBoost & 3.577 / +1.517 & 3.101 / -0.716 & 3.019 / +0.110\\
Elia & GBDT & 3.610 / +1.562 & 3.113 / -0.681 & 3.018 / +0.058\\
Elia & LightGBM & 3.501 / +1.429 & 3.067 / -0.691 & 2.936 / +0.088\\
\bottomrule\end{tabular}\end{table}

\subsection{How recent power contributes}
In the shorter matched fits, adding one residual scale to the weather response reduced nMAE in every period. Four scales reduced it further (Table~\ref{tab:matched}). The four-minus-single differences were $-0.309$ ($[-0.364,-0.252]$), $-0.211$ ($[-0.295,-0.134]$), and $-0.298$ ($[-0.367,-0.230]$) percentage points. Four scales also beat the direct scalar head on overall error by 0.254, 0.625 and 0.340 points. At 240 minutes, each interval for this latter contrast touched or crossed zero at the displayed precision. Changing only the survival law gave a different answer: four-scale exponential minus unrestricted coefficients was $+0.000$, $+0.156$ and $-0.025$ points, with only the second interval excluding zero, in favour of the unrestricted coefficients. The clearest evidence is for keeping several signed residual summaries. Supplementary Section~\ref{Sup-sec:controlleads} gives paired lead-specific intervals.

\begin{table}[tbp]
\centering
\caption{Budget-matched joint fits on 18 public series. Daytime nMAE is expressed as a percentage of capacity. Each row averages three forecasts before scoring. The lower pair uses the complete covariate pathways.}
\label{tab:matched}
\begin{tabular}{@{}lrrr@{}}\toprule
Model & I & II & III\\\midrule
Shared response alone & 7.424 & 7.503 & 8.626\\
Response + one exponential scale & 6.838 & 6.412 & 7.803\\
Response + four exponential scales & 6.529 & 6.201 & 7.505\\
Response + direct scalar residual head & 6.783 & 6.826 & 7.845\\
Four scales, unrestricted coefficients & 6.529 & 6.045 & 7.529\\\midrule
Full CIFT & 5.943 & 5.555 & 6.867\\
Public TSMixer, matched weather and training & 6.086 & 6.374 & 7.371\\\bottomrule
\end{tabular}
\end{table}

With eight epochs, paired CIFT-minus-TSMixer differences were $-0.144$, $-0.819$ and $-0.504$ points; all three intervals excluded zero. These runs used a common input mapping for TSMixer and identical training and selection conditions across the two models. The full-budget ordinary-telemetry results in Table~\ref{tab:overall} show that this gap depends on the training recipe.

\begin{table}[tbp]\centering\footnotesize
\caption{Change in daytime nMAE after adding each component and retraining. Entries give the difference in capacity percentage points and its paired 95\% interval across three-date blocks. Negative values favour the addition. Each fit used eight epochs, three seeds and ordinary telemetry for evaluation.}\label{tab:reviewtwo-paths}\label{tab:dthree-steps}
\begin{tabular}{@{}lccc@{}}\toprule
Addition & Period I & Period II & Period III\\ \midrule
Memory summary & \shortstack{$-0.070$\\[-1pt]{\scriptsize $[-0.112, -0.025]$}} & \shortstack{$-0.137$\\[-1pt]{\scriptsize $[-0.245, -0.049]$}} & \shortstack{$-0.064$\\[-1pt]{\scriptsize $[-0.118, -0.002]$}} \\[2pt]
Clear-sky-index persistence & \shortstack{$-0.541$\\[-1pt]{\scriptsize $[-0.673, -0.413]$}} & \shortstack{$-0.582$\\[-1pt]{\scriptsize $[-0.676, -0.476]$}} & \shortstack{$-0.451$\\[-1pt]{\scriptsize $[-0.608, -0.317]$}} \\[2pt]
Last-innovation persistence & \shortstack{$-0.023$\\[-1pt]{\scriptsize $[-0.036, -0.010]$}} & \shortstack{$-0.017$\\[-1pt]{\scriptsize $[-0.030, -0.003]$}} & \shortstack{$-0.021$\\[-1pt]{\scriptsize $[-0.037, -0.006]$}} \\[2pt]
Adjacent redistribution & \shortstack{$-0.017$\\[-1pt]{\scriptsize $[-0.024, -0.009]$}} & \shortstack{$-0.013$\\[-1pt]{\scriptsize $[-0.019, -0.006]$}} & \shortstack{$-0.023$\\[-1pt]{\scriptsize $[-0.034, -0.010]$}} \\[2pt]
Weather path, equal blend & \shortstack{$0.260$\\[-1pt]{\scriptsize $[0.155, 0.362]$}} & \shortstack{$0.385$\\[-1pt]{\scriptsize $[0.215, 0.585]$}} & \shortstack{$0.172$\\[-1pt]{\scriptsize $[-0.036, 0.384]$}} \\[2pt]
Learned gate & \shortstack{$-0.158$\\[-1pt]{\scriptsize $[-0.205, -0.111]$}} & \shortstack{$-0.179$\\[-1pt]{\scriptsize $[-0.233, -0.126]$}} & \shortstack{$-0.166$\\[-1pt]{\scriptsize $[-0.248, -0.080]$}} \\[2pt]
Auxiliary vintages & \shortstack{$-0.037$\\[-1pt]{\scriptsize $[-0.101, 0.023]$}} & \shortstack{$-0.103$\\[-1pt]{\scriptsize $[-0.172, -0.049]$}} & \shortstack{$-0.085$\\[-1pt]{\scriptsize $[-0.179, 0.008]$}} \\[2pt]
\midrule
Full minus four-scale fit & \shortstack{$-0.586$\\[-1pt]{\scriptsize $[-0.753, -0.410]$}} & \shortstack{$-0.646$\\[-1pt]{\scriptsize $[-0.739, -0.541]$}} & \shortstack{$-0.638$\\[-1pt]{\scriptsize $[-0.805, -0.478]$}} \\[2pt]
\bottomrule\end{tabular}\end{table}

Clear-sky-index persistence gave the largest reduction in all three periods (Table~\ref{tab:reviewtwo-paths}). Adjacent redistribution contributed 0.013--0.023 points. Using $q$ directly as the historical reference raised ordinary error in I and II, with an interval containing zero in III. The complete model improved on the history-only fit with 24 hours of telemetry masked in II and III, and under ordinary telemetry the three paired intervals spanned zero.

\subsection{Does the forecast correction need exponential attenuation?}
Retraining the simplified four-scale control with either forward coefficient rule preserves the form of the historical kernels. Equation~\eqref{eq:memory} has a physical analogy: if the local mismatch relaxes at a constant fractional rate, $de/dt=-\lambda e$ gives exponential attenuation. The fitted rate varies by lead and is learned from the residuals themselves, not derived from measured cloud lifetimes or plant recovery times. A cloud front arriving later, a returning shadow or a sustained power limit need not produce monotone relaxation. The kernel on past observations, the forward multiplier and the final mixture are distinct; even the constrained model's final correction can increase.

The sign also depends on lead. In I the constrained control gained 0.068 points at one hour; in III it gained 0.392 at 15 minutes. In II the constrained form had higher error by 0.094, 0.248 and 0.162 points at one, two and four hours. These are pointwise contrasts (Table~\ref{Sup-tab:structleads}) and were not used to select a lead-specific rule.

We tested the weather-regime interpretation using frozen forecasts. For origins with all 16 targets valid and in daylight, we split the four-hour range of issued hourly cloud cover at 20 percentage points. This exploratory threshold defines forecast variation at grid scale. In period II, exponential minus unrestricted nMAE was $+0.211$ points in the lower-range group (95\% interval $[0.067,0.383]$) and $+0.266$ in the higher-range group ($[0.149,0.383]$). Their paired difference was $+0.055$ ($[-0.091,0.180]$). The corresponding interactions in periods I and III were statistically comparable (Table~\ref{Sup-tab:r4-regime}). The period-II contrast therefore appears in both groups, so cloud-cover range does not separate the two rules. We read the forward decay law as a tunable design choice that the data can select per period rather than as a regime-dependent switch.

\subsection{Telemetry and archived weather}
Section~\ref{Sup-sec:r4-supp-controls} reports replays of the selected checkpoints with delayed or missing power and alternative weather inputs. At four hours in period III, replacing the weather fields alone increased error by 0.573 points; changing only their forecast-lead bin and associated age inputs increased it by 2.149. Because all 1,032,944 training origin--lead pairs used primary bin 1, the second figure indexes sensitivity to an input outside the training support rather than the operational value of a weather update (Section~\ref{Sup-sec:r3weather}).

\section{Discussion}
\subsection{Power history at the first quarter-hour}
At one quarter ahead, the latest measured operating level is often close to the target, and the projected ratio accounts for the change in solar elevation over those 15 minutes. The three tree families receive the last valid power, its recent ratio to the solar envelope, and that ratio projected to target time. Each lead has its own fit. CIFT shares its response across the trajectory and across ordinary and masked histories. These differences in training and representation mean that a lead-specific contrast reflects several design factors jointly.

Restoring the last meter reading and updating the operating scale in frozen CIFT reduced error most at the first lead. In period III the reduction was greater at the Australian plants than in the Belgian provincial totals (Figure~\ref{Sup-fig:jointreplay}). Under ordinary telemetry the two decay rules differed little at the first lead in periods I and II, and the constrained fit had lower first-lead error in III (Table~\ref{Sup-tab:structleads}). The telemetry replay and decay refits distinguish sensitivity to a recent reading from sensitivity to the chosen propagation law.

\subsection{What clear-sky-index persistence captures}
A meter reading taken during the morning rise can remain informative even as its value in megawatts becomes outdated. Carrying the last daylight power-to-envelope ratio forward lets the expected output rise with the target envelope. Where the denominator floor and clipping are inactive, the absolute error of this candidate equals the target envelope times the absolute change in that ratio. The ratio can reflect cloud cover, clipping or soiling. Its usefulness depends on how long the attenuation persists.

In the eight-epoch refits, adding this candidate after the memory summary gave the largest decrease in nMAE along the tested sequence, with paired intervals below zero in all three periods (Table~\ref{tab:reviewtwo-paths}). The 0.21--0.31 point four-versus-one contrast isolates three additional scales, whereas the first scale alone had already reduced error by 0.59--1.09 points against the response (Table~\ref{tab:matched}). The 0.45--0.58 point increment from the solar index is therefore an additive gain on top of the residual path rather than a measure of the path's total value. Its gain persisted when the last-innovation candidate entered first, although its size changed (Appendix~\ref{Sup-sec:dthree-details}). A separate persistence replay held the index and fallback rules fixed. Projecting that index with the same site envelope used to compute it lowered error in every period (Table~\ref{Sup-tab:review2-reference}). At Beryl on 19 October, all three plotted models overestimated mean output during a midday dip (Figure~\ref{Sup-fig:berylcase}). The hourly shortwave input contains no within-hour cloud variation, and without measured plane-of-array irradiance or module temperature a passing cloud and a change in plant operation look the same in the power record.

\subsection{Period II and the limits of transfer}
Elia pools observations from multiple installations to estimate provincial output. Local power drops can partly cancel, whereas a cloud crossing a single plant can produce a sharp change at its meter. The regional contrasts are compatible with this mechanism, although region, season, reporting scale and dispatch policy vary together, so each contrast describes their joint effect rather than residual memory at a fixed spatial scale. The period-II cost of constraining the residual persisted in both cloud-cover groups, and separating passing clouds from these other factors would need cloud-motion observations at the spatial and temporal scale of the power series.

The measured power-to-envelope ratio remained correlated across four hours in many series (Section~\ref{Sup-sec:r8dependence}). Its serial dependence is a different quantity from the signed error of the weather-conditioned response that enters CIFT, so these correlations serve as a comparison scale rather than an estimate of cloud transit time.

\subsection{Evaluation scope}
The evaluation revisits known sites and scores overlapping forecasts, so the paired date-block intervals describe transfer within these series, periods and fits. An isolated dispatch cap can leave high ratios in the preceding 72 hours, so the 95th-percentile operating scale $A_s$ can remain high; a persistent cap can lower that scale and lead to underprediction after release. The flagged targets were excluded, so both responses are evaluated outside curtailment. Australian targets pass a dispatch screen, and the Belgian archive changes inside period III without overlapping observations from both products. The weather replays change the primary fields together with the archive bin, which also enters a bin value outside the training support, and the run-resolved check uses eight dates at three plants. Prospective verification would require forecasts saved at both issue and receipt times for weather and telemetry.

The physical comparator uses one representative plane and nominal ratings. The power-error contrasts therefore describe the combined effect of irradiance correction, conversion efficiency and operating restrictions; separating them would need measured plane-of-array irradiance and a plant-calibrated conversion chain. Training uses night labels, whereas the reported errors condition on daylight. The envelope quantile (0.97), operating quantile (0.95), operating-scale cutoff (0.30) and adjacent-transfer bound (0.45) were held fixed across series, and the evaluation spans mid-July to mid-November rather than a full annual cycle.

\section{Conclusions}
A solar-relative persistence candidate and several signed summaries of recent power discrepancies each improved the tested eight-epoch models. An exponential multiplier serves as an effective default rather than a required form, and the reported increments correspond to matched eight-epoch refits. For the selected full model, the three-seed forecast achieved daylight nMAE of 5.203, 5.062 and 6.336\% of capacity. The three-seed CIFT forecast gained 0.207 capacity percentage points over LightGBM across the full trajectory in period I. Scored per training seed rather than as the three-seed average, the mean period-I advantage over LightGBM is 0.086 points, with a paired interval that spans zero (Supplementary Table~\ref{Sup-tab:individual}); part of the all-lead gain over the strongest tree therefore comes from seed averaging, which belongs to the reported forecast. These temporal replays at known sites define the scope, and a new site or a later operating period is the natural next test.

\section*{Data availability}
Generation records are available from Elia ODS032 and ODS209 and AEMO NEMWeb. Weather fields come from the Open-Meteo Previous Runs and Single Runs APIs \cite{elia,elia209,aemo,openmeteo}. Table~\ref{Sup-tab:ids} and \path{diagnostics/sites.csv} identify the series, dispatch units and weather coordinates.

\section*{Code availability}
The source package contains the neural and tree implementations, three CIFT checkpoints, June selection records and figure data. Issue-level forecasts from each CIFT seed, XGBoost, GBDT, LightGBM, ordinary-telemetry TSMixer and the persistence references share keys with the archived targets and masks. Scripts in \path{diagnostics/} reproduce the comparison at all 16 leads. A small prepared example exercises the training and inference code. A full refit requires prepared training data for all 18 series. Supplementary Section~\ref{Sup-sec:publiccode} links the neural source revisions.

\bibliographystyle{elsarticle-num}
\setlength{\bibsep}{0pt plus 0.2pt}
\bibliography{references}
\end{document}